\documentclass[runningheads]{llncs}
\usepackage{graphicx}
\usepackage{comment}
\usepackage{amsmath,amssymb} %
\usepackage{color}

\usepackage{epsfig}
\usepackage{ctable}
\usepackage{ulem}
\usepackage{color}
\usepackage{multirow} %
\usepackage{cite}
\usepackage{subfig}
\usepackage{algorithm,algpseudocode}

\newcommand{\pnp}{P\&P}

\newcommand{\tabref}{Table.~\ref}
\newcommand{\figref}{Fig.~\ref}
\newcommand{\etal}{\textit{et al.}}

\begin{document}
\pagestyle{headings}
\mainmatter
\def\ECCVSubNumber{3838}  %

\title{V2VNet: Vehicle-to-Vehicle Communication for Joint Perception and Prediction} %

\begin{comment}
\titlerunning{ECCV-20 submission ID \ECCVSubNumber}
\authorrunning{ECCV-20 submission ID \ECCVSubNumber}
\author{Anonymous ECCV submission}
\institute{Paper ID \ECCVSubNumber}
\end{comment}

\titlerunning{V2VNet: V2V Communication for Joint Perception and Prediction}
\author{Tsun-Hsuan Wang\inst{1} \and
Sivabalan Manivasagam\inst{1,2} \and
Ming Liang\inst{1} \and
Bin Yang\inst{1,2} \and
Wenyuan Zeng\inst{1,2} \and
James Tu\inst{1,2} \and
Raquel Urtasun\inst{1,2}
}
\authorrunning{T.H. Wang et al.}
\institute{UberATG \and University of Toronto\\
\email{\{tsunhsuan.wang, manivasagam, ming.liang, \\byang, wenyuan, james.tu, urtasun\}@uber.com}}
\maketitle

\begin{abstract}
	In this paper, we explore the use of vehicle-to-vehicle (V2V) communication to improve the perception and motion forecasting performance of self-driving vehicles.
By intelligently aggregating the information received from multiple nearby vehicles, we can 
observe the same scene from different viewpoints. 
This allows us to see through occlusions and 
detect
 actors at long range, where the observations are very sparse or non-existent. 
We also show that our approach of sending 
compressed deep feature map activations
achieves high accuracy while satisfying communication bandwidth requirements.
\end{abstract}

\section{Introduction}

\begin{figure}[t]
  \centering
  \includegraphics[width=0.27\linewidth]{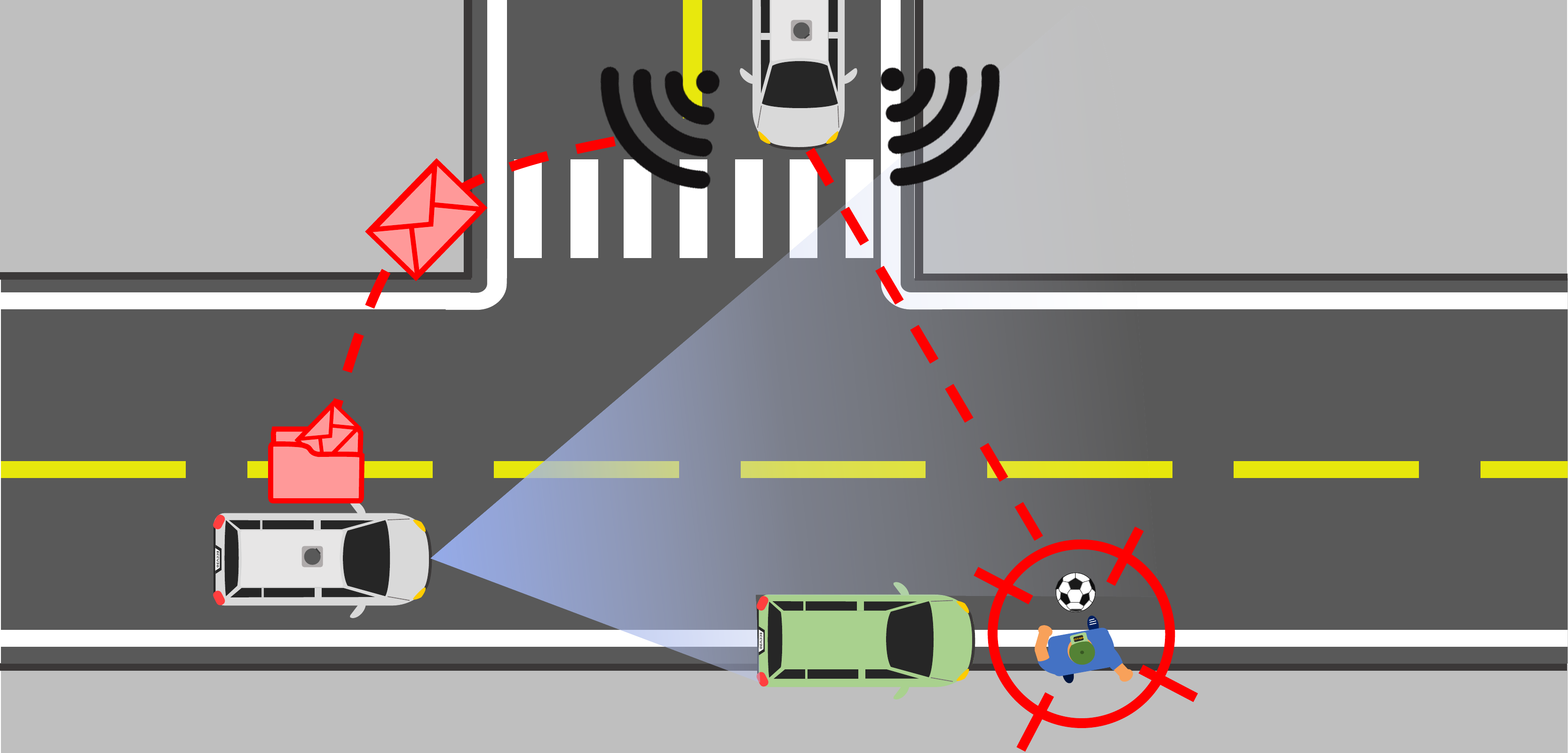}
  \includegraphics[width=0.72\linewidth]{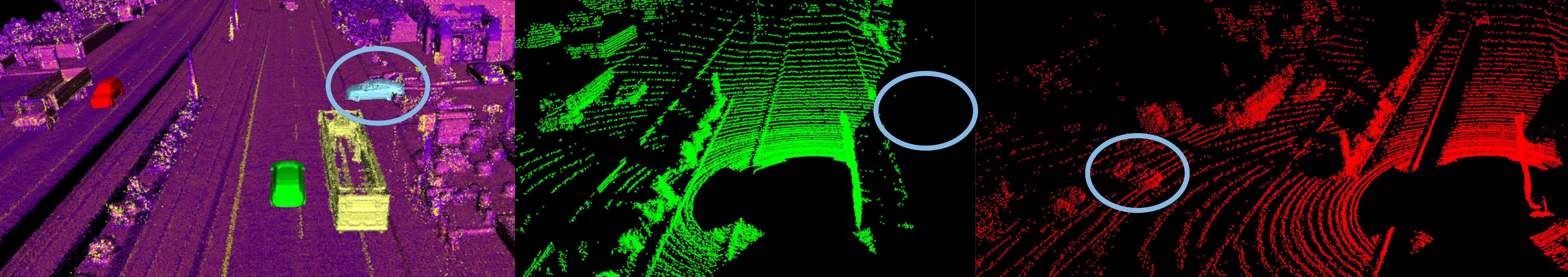}
 \caption{
\textit{Left:} Safety critical scenario of a pedestrian coming out of occlusion. 
V2V communication can be leveraged to use the fact that multiple self-driving vehicles see the scene from different viewpoints, and thus see through occluders.
\textit{Right:} 
Example \textit{V2VSim} Scene. Virtual scene with occluded actor (blue) and SDVs (red and green), Rendered LiDAR from each SDV in the scene.
}
  \label{fig:teaser}
\end{figure}

While a world densely populated with self-driving vehicles (SDVs) might seem futuristic, these vehicles will one day soon be the norm.
They will provide safer, cheaper and less congested transportation solutions for everyone, everywhere. 
A core component
of self-driving vehicles is their ability to perceive the world. 
From sensor data, 
the SDV 
needs to reason about the scene in 3D, identify the other agents, 
and forecast how their futures might play out.
These tasks are commonly referred to as perception and motion forecasting.
Both strong perception and motion forecasting  are critical for the SDV to plan and maneuver through traffic to get from one point to another safely.

The reliability of perception and 
motion forecasting
algorithms has significantly improved in the past few years due to the development of neural network architectures that can reason in 3D and intelligently fuse multi-sensor data (e.g., images, LiDAR, maps)  \cite{liang2018deep, liang2019multi}.
Motion forecasting algorithm performance has been further improved by 
building good multimodal distributions  \cite{jain2019discrete, chai2019multipath, cui2019deep,ilvm} that capture diverse actor behaviour
and by modelling actor interactions \cite{casas2019spatially, rhinehart2018r2p2, rhinehart2019precog,li2020end}.
Recently, 
\cite{luo2018fast, casas2018intentnet} propose 
approaches that perform joint perception and 
motion forecasting, dubbed \textit{perception and prediction} (\pnp),
further increasing the accuracy while being computationally more efficient than classical two-step pipelines.

Despite these advances,  
challenges 
remain.
For example, objects that are heavily occluded or far away
result in 
sparse observations and pose a challenge for modern computer vision systems.
Failing to detect and predict the intention of these hard-to-see actors might have catastrophic consequences in safety critical situations when there are only a few miliseconds to react: 
imagine the SDV driving along a road and a 
child chasing after a soccer ball runs into the street from behind a parked car (Fig. \ref{fig:teaser}, left). 
This situation is 
difficult for both SDVs 
and
human drivers to correctly perceive and adjust for. 
The crux of the problem is that 
the SDV and the human can only see the scene from a single viewpoint. 

However,  SDVs 
could
have super-human capabilities if we equip them with the ability to transmit information and utilize the information received from nearby vehicles to better perceive the world.   
Then the SDV could
see behind the 
occlusion and detect the 
child 
earlier, allowing for a safer 
avoidance maneuver. 

In this paper, we consider the \textit{vehicle-to-vehicle} (V2V) 
communication 
setting, where each vehicle can broadcast and receive information
to/from nearby vehicles (within a 70m radius). 
Note that this broadcast range is realistic based on existing communication protocols \cite{kenney2011dsrc}. 
We show that to achieve the best compromise of having strong perception and motion forecasting performance while also satisfying existing hardware transmission bandwidth capabilities, we should send compressed intermediate representations of the \pnp { } neural network. 
Thus, we derive a novel \pnp {} model, called {\it V2VNet}, which utilizes a spatially aware graph neural network (GNN) to aggregate the information received from all the nearby SDVs,  allowing us to intelligently combine information from different points in time and viewpoints in the scene. 

To evaluate our approach, we require a dataset where multiple self-driving vehicles are in the same local traffic scene. Unfortunately, no such dataset exists.
Therefore, our second contribution is a new dataset, dubbed {\it V2V-Sim} (see Fig. \ref{fig:teaser}, right) that mimics the setting where there are multiple SDVs driving in the area.
Towards this goal, we use a high-fidelity LiDAR simulator \cite{siva2019lidarsim}, which uses a large catalog of  static 3D scenes and dynamic objects built from real-world data,
to simulate realistic LiDAR point clouds for a given traffic scene.
With this simulator, we can recreate traffic scenarios recorded from the real-world and simulate them as if a percentage of the vehicles are SDVs in the network.
We show that V2VNet and other V2V methods significantly boosts 
performance relative to the single vehicle system, and that our compressed intermediate representations reduce bandwidth requirements without sacrificing performance.
We hope this work brings attention to the potential benefits of the V2V setting 
for bringing safer autonomous vehicles on the road. 
To enable this, we plan to release this new dataset and make a challenge with a leaderboard and evaluation server.

\begin{figure*}[t]
  \centering
  \includegraphics[width=\linewidth]{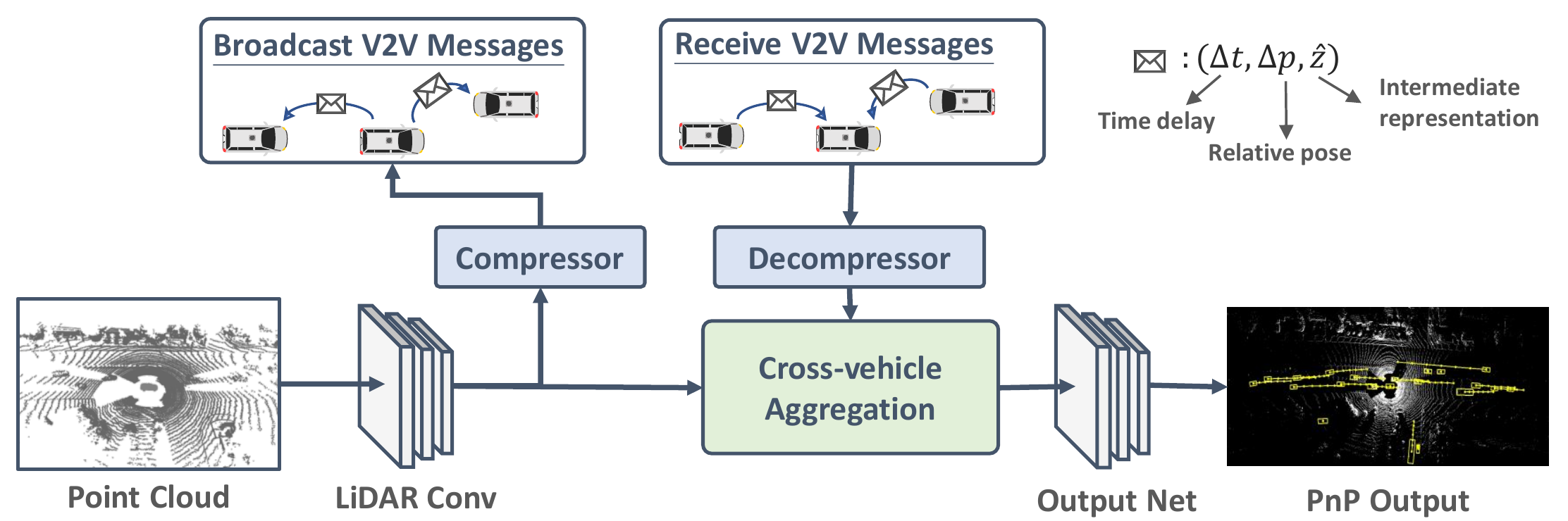}
  \caption{Overview of V2VNet.}
  \label{fig:overview}
 \end{figure*}
 
\section{Related Work}

\subsubsection{Joint Perception and Prediction:}
\label{sec:pnp_related}
Detection and motion forecasting play a crucial role in any autonomous driving system.
\cite{luo2018fast, casas2018intentnet,casas2019spatially,ilvm,pnpnet,li2020end} unified 3D detection and motion forecasting for self-driving, gaining  two key benefits:
(1) Sharing computation of both tasks achieves efficient memory usage and fast inference time.
(2) Jointly reasoning about detection and motion forecasting improves accuracy and robustness.
We build upon these existing \pnp{ }
models by  incorporating V2V communication to share information from different SDVs,  enhancing detection and motion forecasting.

\subsubsection{Vehicle-to-Vehicle Perception:}
For the perception task, prior work has utilized messages encoding three types of data: raw sensor data, output detections, or metadata messages that contain vehicle information such as location, heading and speed.
\cite{rockl2008v2v, rauch2012car2x} associate the received V2V messages with outputs of local sensors.
\cite{chen2019cooper} aggregate LiDAR point clouds from other vehicles, followed by a deep network for detection.
\cite{xiao2018multimedia, rawashdeh2018collaborative} process sensor measurements via a deep network and then generate perception outputs for cross-vehicle data sharing.
In contrast, we
leverage the
power of deep networks by transmitting a compressed intermediate representation.
Furthermore, while previous works demonstrate results on a limited number of simple and unrealistic scenarios, we showcase the effectiveness of our model on a diverse large-scale self-driving V2V  dataset.

\begin{figure*}[t]
  \centering
  \includegraphics[width=\linewidth]{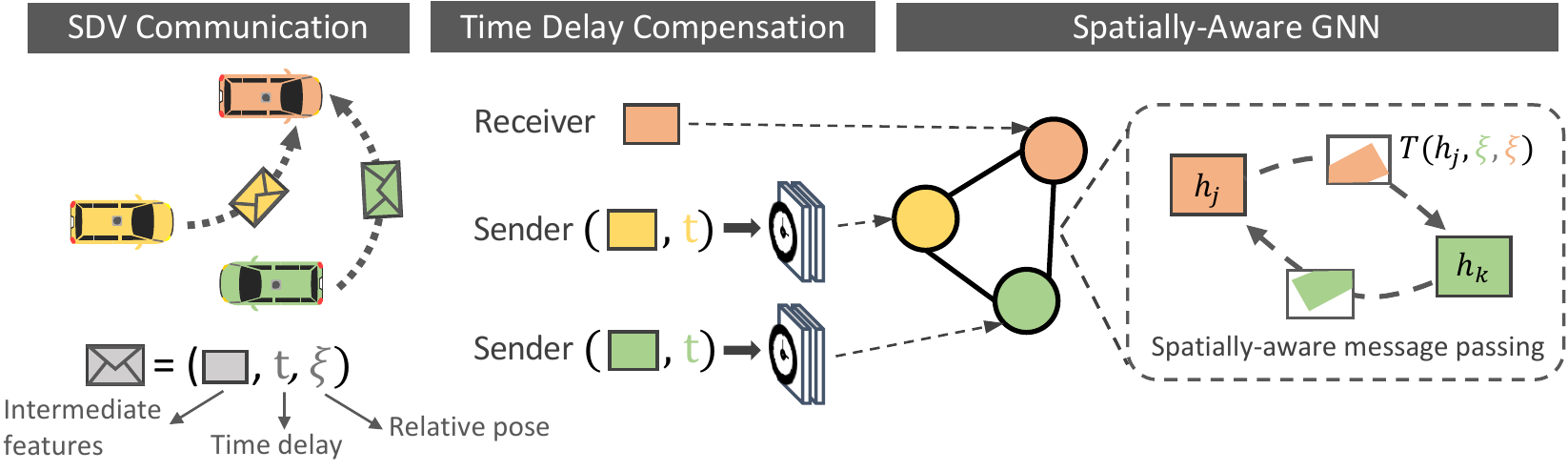}
  \caption{After SDVs communicate  messages, each receiver SDV compensates for time-delay of the received messages, and a GNN aggregates the spatial messages to compute the final intermediate representation.%
}
  \label{fig:gnn}
\end{figure*}

\subsubsection{Aggregation of Multiple Beliefs:}
\label{sec:gnn_related}
 In V2V setting, the receiver vehicle should
collect and aggregate information from an arbitrary number of sender vehicles for downstream inference.
A straightforward approach is to perform permutation invariant operations such as pooling \cite{su2015multi, chen2017multi} over features from different vehicles. 
However, this strategy ignores cross-vehicle relations (spatial locations, headings, times)
and fails to jointly reason about features from the sender and receiver.
On the other hand, recent work on graph neural networks (GNNs) has shown 
success on processing graph-structured data \cite{duvenaud2015convolutional, hamilton2017inductive, li2017situation, yu2017spatio}.
MPNN \cite{gilmer2017neural} abstract commonalities of GNNs with a message passing framework.
GGNN \cite{li2015gated} introduce a gating mechanism for node update in the propagation step.
Graph-neural networks have also be effective in self-driving: 
\cite{casas2019spatially,li2020end} propose a spatially-aware GNN and an interaction transformer to model the interactions between actors in self-driving scenes. \cite{sykora2020multiagent} uses GNNs to estimate value functions of map nodes and share vehicle information for coordinated route planning.  
We believe GNNs are 
tailored for V2V communication, as each vehicle can be a node in the graph.
V2VNet leverages GNNs to aggregate and combine messages from other vehicles.

\subsubsection{Active Perception:}
\label{sec:active_perception}
  In V2V perception, the receiving vehicle should aggregate information from different viewpoints such that its field of view is maximized, trusting more the view that can see better. 
Our work is related to a long line of work in active perception, which focuses on deciding what action the agent should take to better perceive the environment. 
Active perception has been effective in localization and mapping \cite{davison2002simultaneous, kim2015active}, vision-based navigation \cite{davison1999mobile}, serving as a learning signal \cite{jayaraman2016look,yun2017action}, and various other robotics applications \cite{chen2011active}. 
In this work, rather than actively steering SDVs to obtain better viewpoint and sending information to the others, we consider a more realistic scenario where multiple SDVs have their own routes but are currently in the same geographical area, allowing the SDVs to see better by sharing perception messages.

\section{Perceiving the World by Leveraging Multiple Vehicles}

In this paper, we design a novel perception and motion forecasting  model that enables the self-driving vehicle to leverage the fact that several SDVs may be present in the same geographic area.
Following the success of joint perception and prediction
algorithms \cite{luo2018fast, casas2018intentnet, casas2019spatially, pnpnet}, which we call \pnp, we design our approach as a
joint
architecture to perform both tasks, which is enhanced to
incorporate information received from other vehicles.
Specifically, we would like to devise our \pnp{ } model to do the following: given sensor data
the SDV should (1) process this data, (2) broadcast it, (3) incorporate information received from other nearby SDVs, and then (4) generate final estimates of where all traffic participants are in the 3D space and their predicted future trajectories.

Two key questions arise in the V2V setting:
(i) what information should each vehicle broadcast to retain all the important information while minimizing the transmission bandwidth required?
(ii) how should each vehicle incorporate the information received from other vehicles to increase the accuracy of its perception and motion forecasting outputs?
In this section we address these two questions.

\subsection{Which Information should be Transmitted}
\label{sec:type_of_data}
An SDV can choose to broadcast three types of information:
(i) the raw sensor data, (ii) the intermediate representations of its \pnp{}
system, or (iii) the output detections and motion forecast trajectories.
While all three message types are valuable for improving performance, we would like to minimize the message sizes while maximizing
\pnp{} 
accuracy gains.
Note that small message sizes are critical because we want to leverage cheap, low-bandwidth, decentralized communication devices.
While sending raw measurements minimizes information loss, they require more bandwidth.
Furthermore, the receiving vehicle would need to process all additional 
sensor data received, which might prevent it from meeting the real-time inference requirements.
On the other hand, transmitting the outputs of the
\pnp{}
system is very good in terms of bandwidth, as only a few numbers need to be broadcasted.
However,  we may lose valuable scene context and uncertainty information that could be very important to better  fuse the information.

In this paper, we argue that sending intermediate representations of the
\pnp{}
network achieves the best of both worlds.
First, each vehicle processes its own sensor data and computes its intermediate feature representation. 
This is compressed and broadcasted to nearby SDVs.
Then, 
each SDV's
intermediate representation is updated 
using
the received messages from other SDVs.
This 
is further processed through additional network layers to produce the final perception and motion forecasting
outputs.
This approach has two 
advantages:
(1) Intermediate representations in deep networks can be easily compressed \cite{choi2018high,Wei_2019_CVPR}, while retaining important information for downstream tasks.
(2) It has low computation overhead, as the 
sensor data from other vehicles has already been pre-processed.

\begin{algorithm}[t]
\caption{Cross-vehicle Aggregation}
\label{alg:cva}
\begin{algorithmic}[1]
\State \textbf{input:} 
representation $\hat{z}_i$,  relative pose $\Delta p_i$, and time delay $\Delta t_{i \rightarrow k}$ for each SDV $i$ %
\For{each vehicle $i$}
\State $h_{i}^{(0)} = CNN(\hat{z}_i, \Delta t_{i \rightarrow k}) \mathbin\Vert \mathbf{0}$ \Comment{Compensate time delay, init. node state} \label{cva:init}
\EndFor
\For{$l$ iterations} \Comment{Message passing}
\For{each vehicle $i$} \Comment{Processed in parallel}
    \State $m_{i \rightarrow k}^{(l)} = CNN(T(h_{i}^{(l)}, \xi_{i \rightarrow k}), h_{k}^{(l)})  \cdot M_{i \rightarrow k}$ \Comment{Spatially transform message} \label{cva:spatial}
    \State $h_{i}^{(l+1)} = ConvGRU(h_{i}^{(l)}, \phi_{M}([\forall_{j \in N(i)}, m_{j \rightarrow i}^{(l)}]))$ \Comment{Node state update} \label{cva:update}
\EndFor
\EndFor
\State ${z}^{(L)}_i = MLP(h^{(L)}_{i})$ \Comment{Output updated intermediate representation} \label{cva:final}
\end{algorithmic}
\end{algorithm}

In the following, we first showcase how to compute the intermediate representations and how to compress them.
We then show how  each vehicle should incorporate the received information to increase the accuracy of its
\pnp{}
outputs.

\subsection{Leveraging Multiple Vehicles}
\label{sec:v2vnet}

 V2VNet  has three main stages:
(1) a convolutional network block
 that processes raw sensor 
data and creates a compressible intermediate representation,
(2) a cross-vehicle aggregation stage, which aggregates information received from multiple vehicles with the vehicle's internal state (computed from its own sensor data) to compute an updated intermediate representation,
(3) an output network that computes the final
\pnp{}
outputs.
We now describe these steps in more details.
We refer the reader to \figref{fig:overview} for our 
V2VNet 
architecture.

{\flushleft\bf LiDAR Convolution Block:}
\label{sec:preprocess}
Following the architecture from \cite{yang2018pixor}, we  extract 
features from 
LiDAR data
and transform them into bird's-eye-view (BEV).
Specifically, we voxelize the past five LiDAR point cloud sweeps into
15.6cm$^3$ voxels, apply several convolutional layers, and output feature maps of shape $H\times W\times C$, where $H\times W$ denotes the scene range in BEV, and $C$ is the number of feature channels.
We use 3 layers of 3x3 convolution filters (with strides of 2, 1, 2) to produce a 4x downsampled spatial feature map.
This is the intermediate representation that we 
then compress and broadcast to other nearby SDVs. 

{\flushleft\bf Compression:}
We now describe how each vehicle compresses its intermediate representations prior to transmission.
We adapt
Ball\'e  \etal's variational image compression algorithm \cite{balle2018variational} to compress our intermediate representations; 
a convolutional network learns
to compress our representations with the help of a learned hyperprior. The latent representation is then quantized and 
encoded losslessly with very few bits via entropy encoding.
Note that our compression module is differentiable and therefore trainable, allowing our approach to learn how to preserve the feature map information while minimizing bandwidth.

\begin{table}[t]
  \centering
  \begin{tabular}{l | c  c | c c c |  c}
    \hline
    \multirow{2}{*}{Method} & \multicolumn{2}{c|}{AP@IoU $\uparrow$}
                            & \multicolumn{3}{c|}{$\ell_2$ Error (m) $\downarrow$}
                            & \multicolumn{1}{c}{TCR $\downarrow$} \\
                  & 0.5 & 0.7 & 1.0s & 2.0s & 3.0s & $\tau=$0.01 \\ \hline
    No Fusion     & 77.3 & 68.5 & 0.43 & 0.67 & 0.98 & 2.84  \\
    Output Fusion  & 90.8 & 86.3 & \textbf{0.29} & \textbf{0.50} & 0.80 & 3.00 \\    
    LiDAR Fusion  & 92.2 & 88.5 & \textbf{0.29} & \textbf{0.50} & 0.79 & 2.31  \\
    V2VNet         & \textbf{93.1} & \textbf{89.9} &\textbf{0.29} & \textbf{0.50} & \textbf{0.78} & \textbf{2.25} \\
    \hline
  \end{tabular}
  \caption{Detection Average Precision (AP) at IoU=\{0.5, 0.7\}, prediction with $\ell_2$ error at recall 0.9 at different timestamps, and Trajectory Collision Rate (TCR).}
  \label{tab:general}
\end{table}

{\flushleft\bf Cross-vehicle Aggregation:}
After the SDV computes its intermediate representation and transmits
its compressed bitstream,
it decodes the representation received from other vehicles.
Specifically, we apply entropy decoding to the bit stream and apply a decoder CNN to extract the decompressed feature map.
We then aggregate the received information from other vehicles to
produce an updated intermediate representation.
Our aggregation module has to handle the fact that different SDVs are located at different spatial locations and see the actors at different timestamps due to the rolling shutter of the LiDAR sensor and the different triggering per vehicle of the sensors.
This is important as the intermediate feature representations are spatially aware.

Towards this goal, each vehicle uses a fully-connected graph neural network (GNN) \cite{schlichtkrull2018modeling} as the aggregation module,
where each node in the GNN is the state representation of an SDV in the scene, including itself (see Fig. \ref{fig:gnn}).
Each SDV maintains its own local graph based on which SDVs are within range (i.e., 70 m). GNNs are a natural choice as they handle dynamic graph topologies, which arise in the V2V setting. 
GNNs are deep-learning models tailored to graph-structured data: each node maintains a state representation, and for a fix number of iterations, messages are sent between nodes and the node states are updated based on the aggregated received information using a neural network.
Note that the GNN messages 
are different from
the messages transmitted/received by the SDVs: the GNN computation is done locally by the SDV.
We design our GNN to temporally warp and spatially transform the received messages to the receiver's coordinate system.
We now describe the aggregation process that 
the receiving vehicle performs. %
We  refer the reader to  Alg. \ref{alg:cva} for  pseudocode. %

We first compensate for the time delay between the vehicles to create an initial state for each node in the graph.
Specifically,  for each node, we apply a convolutional neural network (CNN) that takes as input the received intermediate representation $\hat{z}_i$, the relative 6DoF pose  $\Delta p_i$ between the receiving and transmitting SDVs  and the time delay $\Delta t_{i \rightarrow k}$ with respect to the receiving vehicle sensor time. Note that for the node representing the receiving car, $\hat{z}$ is directly its intermediate representation. 
The time delay is computed as the time difference between the sweep start times of each vehicle, based on universal GPS time.
We then take the time-delay-compensated representation and concatenate with zeros to augment the capacity of the node state
to aggregate the information received from other vehicles after propagation (line \ref{cva:init} in Alg. \ref{alg:cva}).

Next we perform GNN message passing.
The key insight is that because the other SDVs are in the same local area, the node representations will have overlapping fields of view.
If we intelligently transform the representations and share information between nodes where the fields-of-view overlap, we can enhance the SDV's understanding of the scene and produce better output \pnp. Fig. \ref{fig:gnn} visually depicts our spatial aggregation module. 
We first apply a relative spatial transformation $\xi_{i \rightarrow k}$ to warp the intermediate state of the $i$-th  node
to send a GNN message to the $k$-th node.
We then perform joint reasoning on the spatially-aligned feature maps of both nodes using a CNN.
The final modified message is computed as in Alg. \ref{alg:cva} line \ref{cva:spatial}, 
where $T$ applies the spatial transformation and resampling of the feature state via bilinear-interpolation, and  $M_{i\rightarrow k}$ masks out non-overlapping areas between the fields of view.
Note that with this design, our messages maintain the spatial awareness.
\begin{figure}[t]
  \centering
  \includegraphics[width=1.0\linewidth]{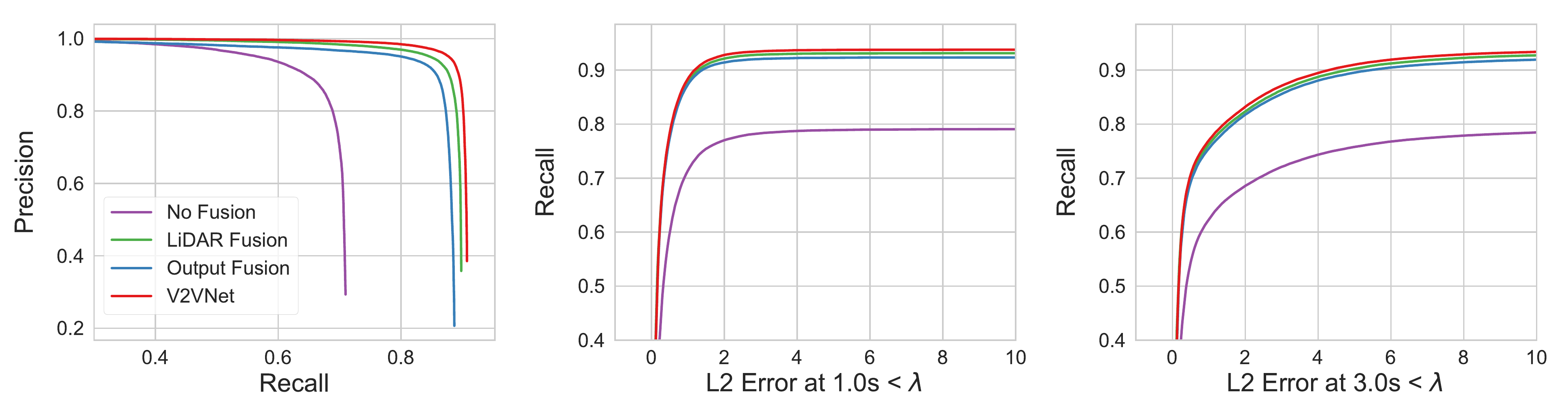}
  \caption{Left: Detection Precision-Recall (PR) Curve at IoU=0.7. Center/Right: Recall as a function of $L_2$ Error Prediction at 1.0s and 3.0s.}
  \label{fig:pr_curve_and_l2_lambda}
\end{figure}

We next aggregate at each node the received messages via a mask-aware permutation-invariant function $\phi_{M}$
and update the node state with a convolutional gated recurrent unit (ConvGRU) (Alg. \ref{alg:cva} line \ref{cva:update}), %
where $j \in N(i)$ are the neighboring nodes in the network for node $i$ and $\phi_{M}$ is the mean operator. %
The mask-aware accumulation operator ensures only
overlapping fields-of-view
are considered.
In addition, the gating mechanism in the node update enables information selection for the accumulated received messages based on the current belief of the receiving SDV.
After the final iteration, a multilayer perceptron outputs the updated intermediate representation (Alg. \ref{alg:cva}  Line \ref{cva:final}). %
We repeat this message propagation scheme for a fix number of iterations.
{\flushleft\bf Output Network:}
\label{sec:output}
After performing message passing, we apply a set of four Inception-like \cite{szegedy2015going} convolutional blocks  %
 to capture multi-scale context  efficiently, which is important for prediction.
Finally, we take the feature map and exploit two network branches to output detection and motion forecasting estimates respectively. The detection output  is  $(x, y, w, h, \theta)$,  denoting the position, size and orientation of each object. The output of the motion forecast branch is parameterized as ${(x_t, y_t)}$, which denotes the object's location at future time step $t$.
We forecast the motion of the actors for the next 3 seconds at 0.5 s intervals. 
Please see supplementary for additional architecture  and implementation details.

\begin{figure}[t]
  \centering
  \includegraphics[width=0.3\linewidth]{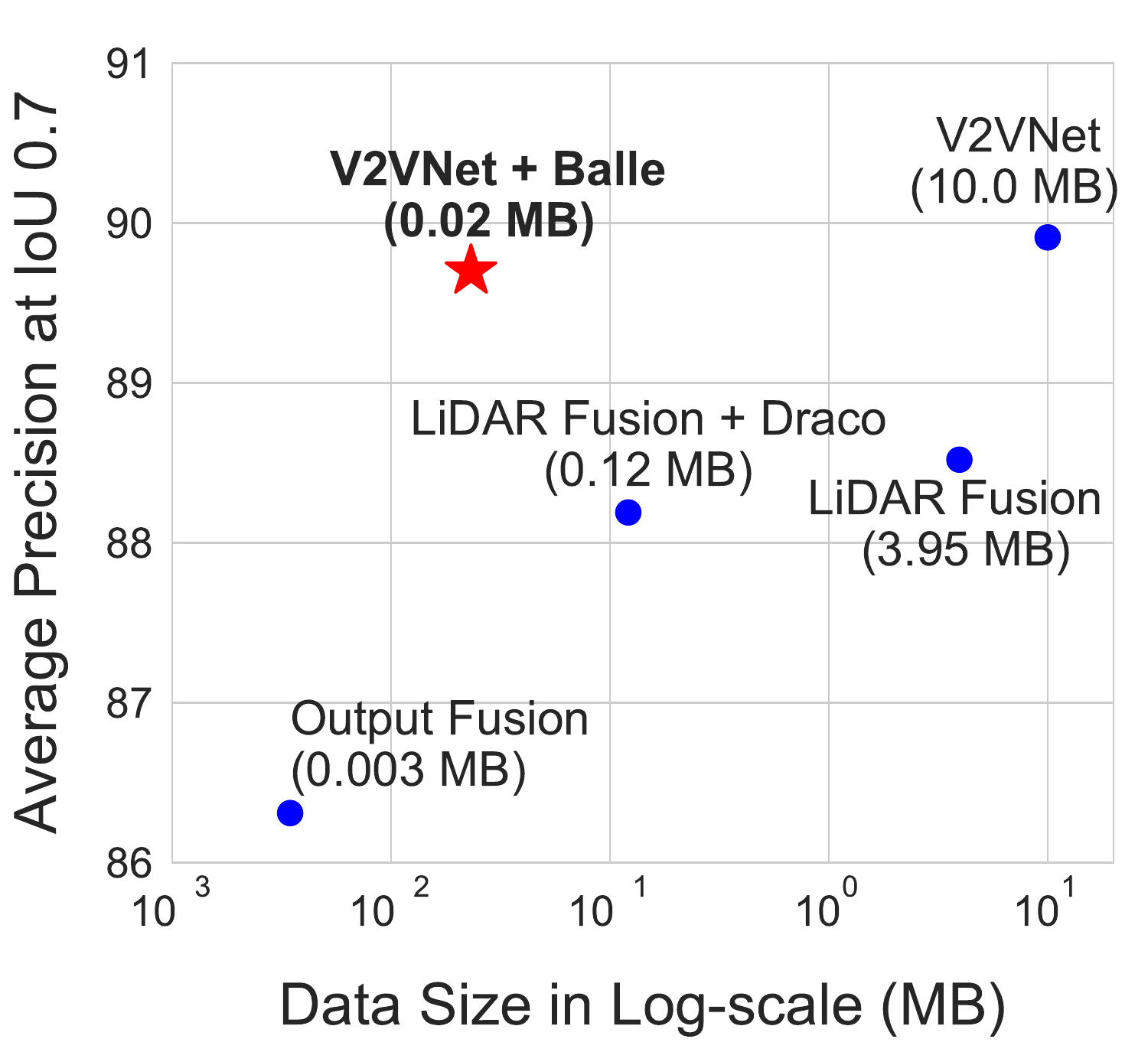}
  \hspace{0.01\linewidth}
  \includegraphics[width=0.3\linewidth]{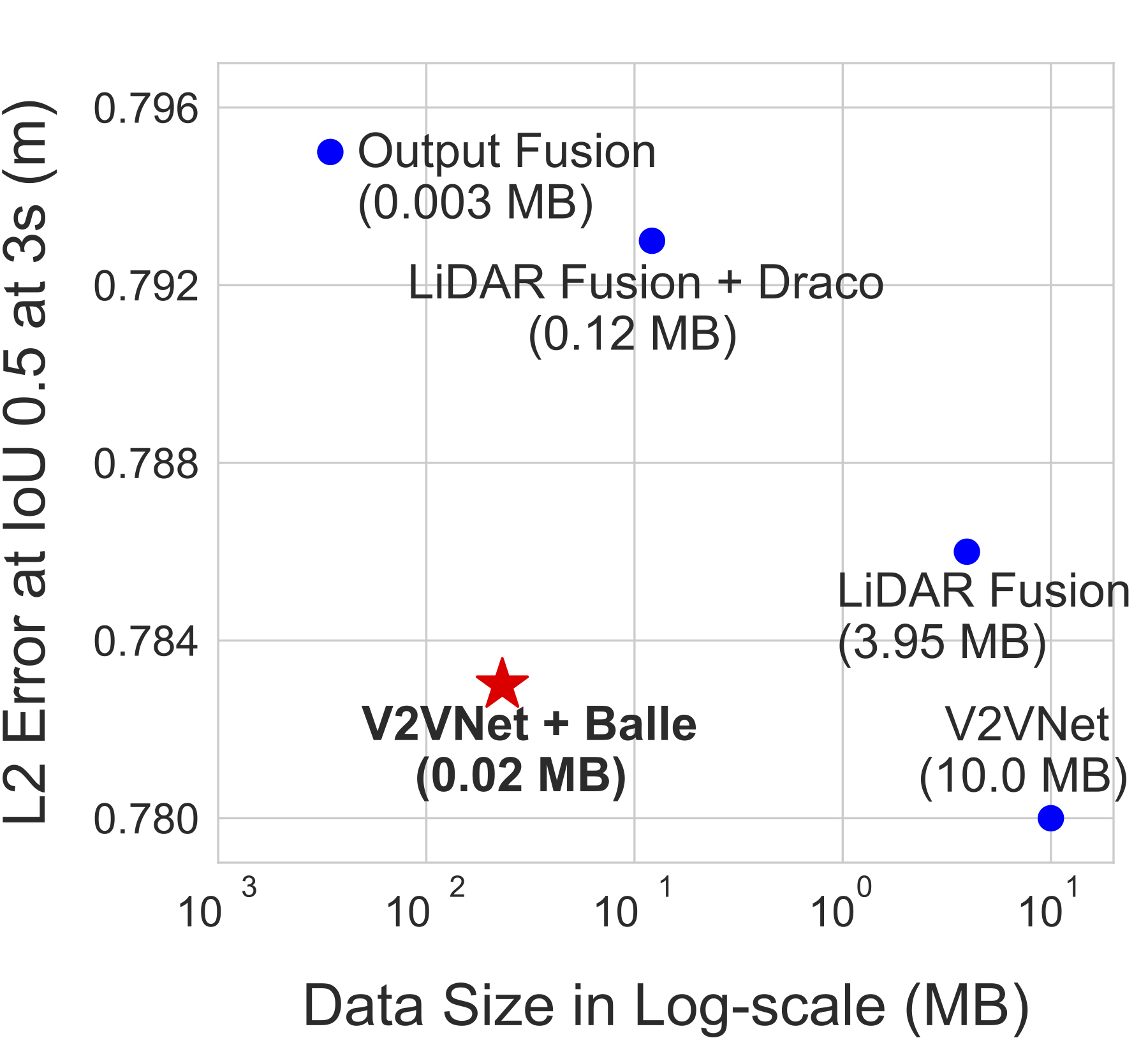}
  \hspace{0.01\linewidth}
  \includegraphics[width=0.3\linewidth]{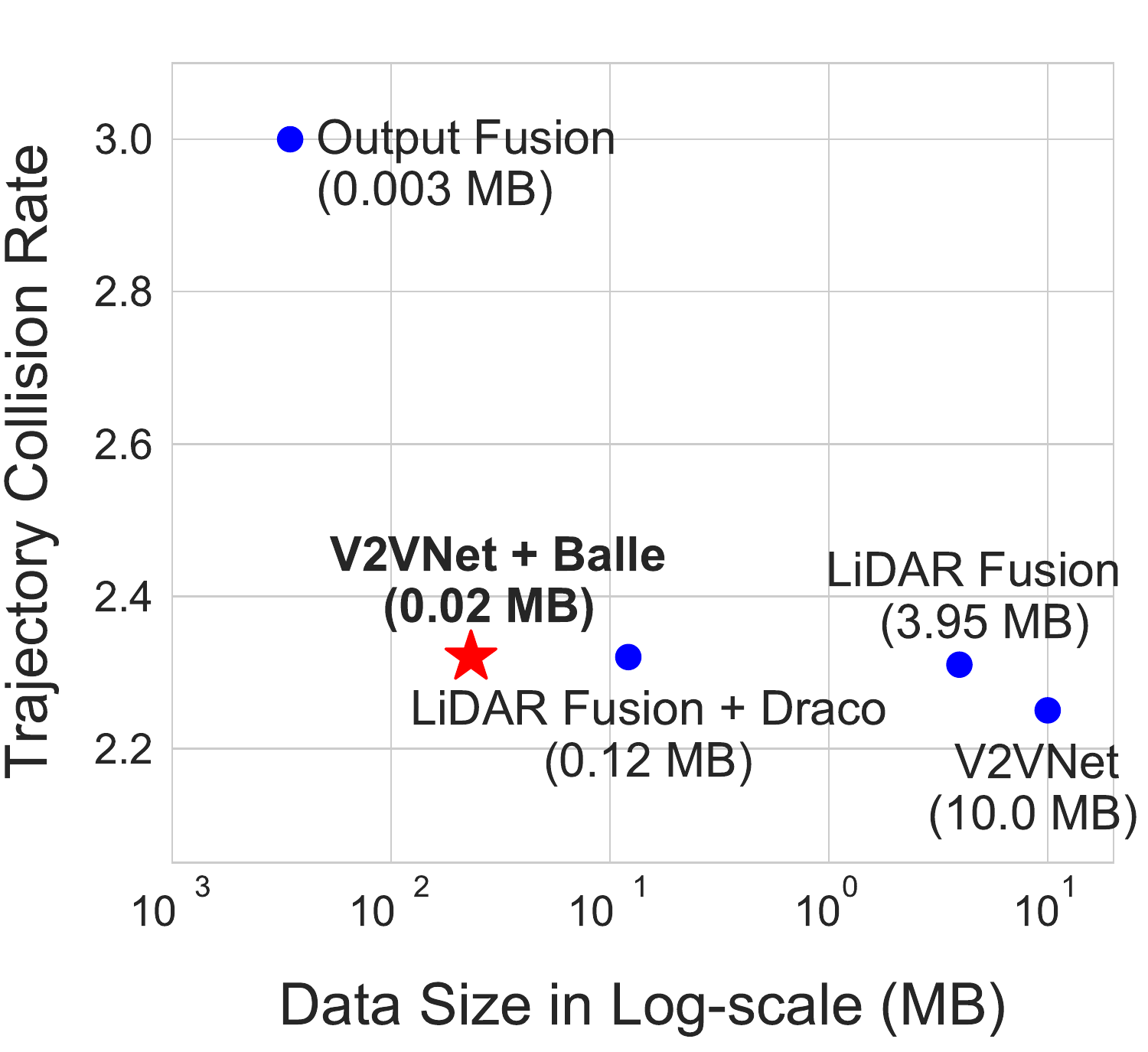}
  
  \caption{{\bf Compression:} Detection (AP at IoU 0.7), Prediction ($\ell_2$ error at recall 0.9 at 3.0s), and Trajectory Collision Rate ($\tau =0.01$) performance on models with compression module. }
  \label{fig:compression}
\end{figure}

\subsection{Learning}
\label{sec:learning}
We first pretrain the LiDAR backbone and output headers,
 bypassing the cross-vehicle aggregation stage.
Our loss function is cross-entropy on the vehicle classification output and smooth $\ell_1$ on the bounding box parameters.
We apply hard-negative mining  to improve performance.
We then  finetune jointly the LiDAR backbone, cross-vehicle aggregation, and output header modules on our novel V2V dataset (see Sec. \ref{sec:dataset}) with synchronized inputs (no time delay) using the same loss function.
We do not use the temporal warping function at this stage.
During training, for every example in the mini-batch, we randomly sample the number of connected vehicles uniformly on $[0, min(c, 6)]$, where $c$ is the number of candidate vehicles available.
This is to make sure V2VNet can handle arbitrary graph connectivity while also making sure the fraction of vehicles 
on the V2V network remains within the GPU memory constraints.
Finally, the temporal warping function is trained to compensate for time delay with asynchronous inputs, where all other parts of the network are fixed.
We uniformly sample time delay between 0.0s and 0.1s (time of one 10Hz LiDAR sweep).
We then train the compression module with the main network (backbone, aggregation, output header) fixed. %
We use a rate-distortion objective, which aims to maximize the bit rate in transmission while minimizing the distortion between uncompressed and decompressed data.
We define the rate objective as the entropy of the transmitted code, and the distortion objective  as the reconstruction loss (between the decompressed and uncompressed feature maps).

\section{V2V-Sim: a dataset for V2V communication}
\label{sec:dataset}

\begin{figure}[t]
  \centering
  \includegraphics[width=1.0\linewidth]{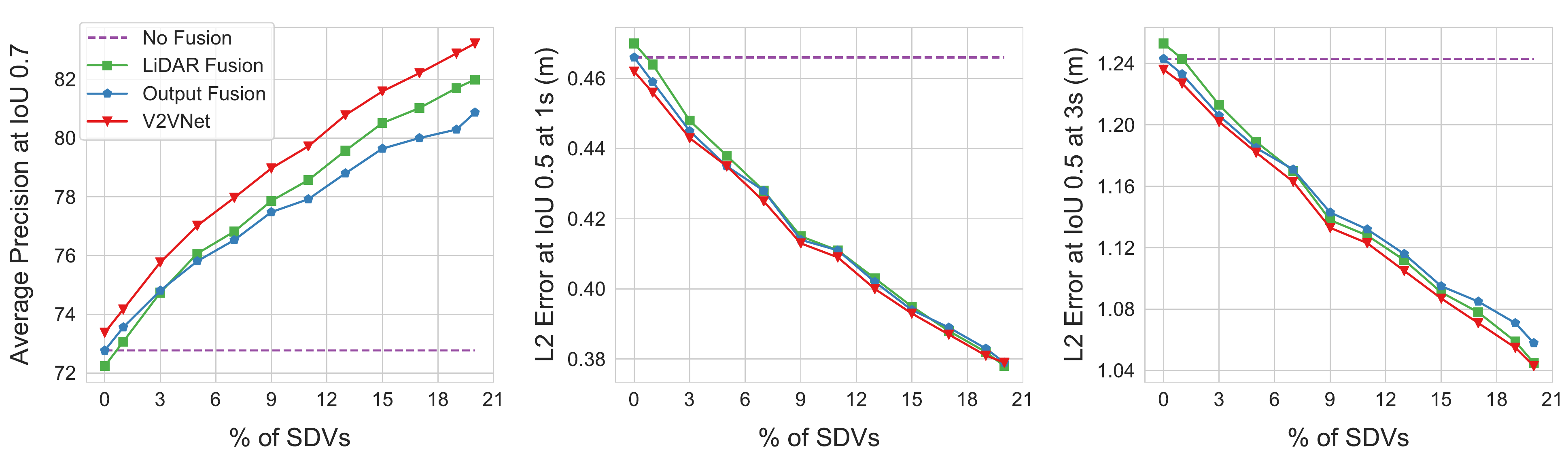}
  \caption{{\bf Density of SDV}: AP at IoU=0.7 and $\ell_2$ error at \{1s, 3s\} at highest recall rate at IoU=0.5 wrt  \% of SDVs in the scene.}
  \label{fig:num_veh}
\end{figure}

No realistic dataset for V2V communication exists in the literature.
Some approaches simulate the V2V setting by using different frames from KITTI \cite{geiger2012we} to emulate multiple vehicles \cite{maalej2017vanets,xiao2018multimedia,chen2019cooper}.
However, this is 
unrealistic 
since sensor measurements are at different timestamps, 
so moving objects may be at completely different locations (e.g., a 1 sec. time difference can cause 20 m change in position).
Other approaches utilize a platoon
strategy for data collection \cite{chen2015dsrc,kim2014multivehicle,yuan2017object,rawashdeh2018collaborative}, where each vehicle  follows behind the previous one closely.
While more realistic than using KITTI,  this data collection is biased: the perspectives of different vehicles are highly correlated with each other, and the data does not provide the richness of different V2V scenarios.
For example, we will never see SDVs  coming in the opposite direction, or SDVs
turning from other lanes at intersections.

To address these deficiencies, we use a high-fidelity LiDAR simulator, LiDARsim \cite{siva2019lidarsim}, to generate our large-scale V2V communication dataset, which we call {\it V2V-Sim}. LiDARsim is a simulation system that uses a large catalog of 3D static scenes and dynamic objects that are built 
from
real-world data collections 
to simulate new scenarios. 
Given a scenario (i.e., scene, vehicle assets and their trajectories), LiDARsim  applies raycasting followed by  a deep neural network to generate a realistic LIDAR point cloud for each frame in the scenario.

We leverage traffic scenarios captured in the real world ATG4D dataset \cite{yang2018pixor} to generate our simulations. %
We recreate the snippets in LiDARsim's virtual world using the ground-truth 3D tracks provided in ATG4D. By using the same scenario layouts and agent trajectories recorded from the real world, we 
can
replicate realistic traffic.
In particular, at each timestep, we place the actor 3D-assets into the virtual scene according to the real-world labels and generate the simulated LiDAR point cloud seen from the different candidate vehicles (see Fig. \ref{fig:teaser}, right). 
We define the candidate vehicles to be non-parked vehicles that are within the 70-meter broadcast range of the vehicle that recorded the real-world snippet. 
We generate 5500 25s snippets collected from multiple cities. We subsample the frames in the snippets to produce our final  46,796/4,404 frames for train/test splits for the V2V-Sim dataset.
V2V-Sim has on average 10 candidate vehicles that could be in the V2V network per sample, with a maximum of 63 and a variance of 7, demonstrating the traffic diversity. 
The fraction of vehicles that are 
candidates 
increases linearly w.r.t broadcast range.

\section{Experimental Evaluation}

\begin{figure*}[t]
  \centering
  \subfloat{\includegraphics[width=0.495\linewidth]{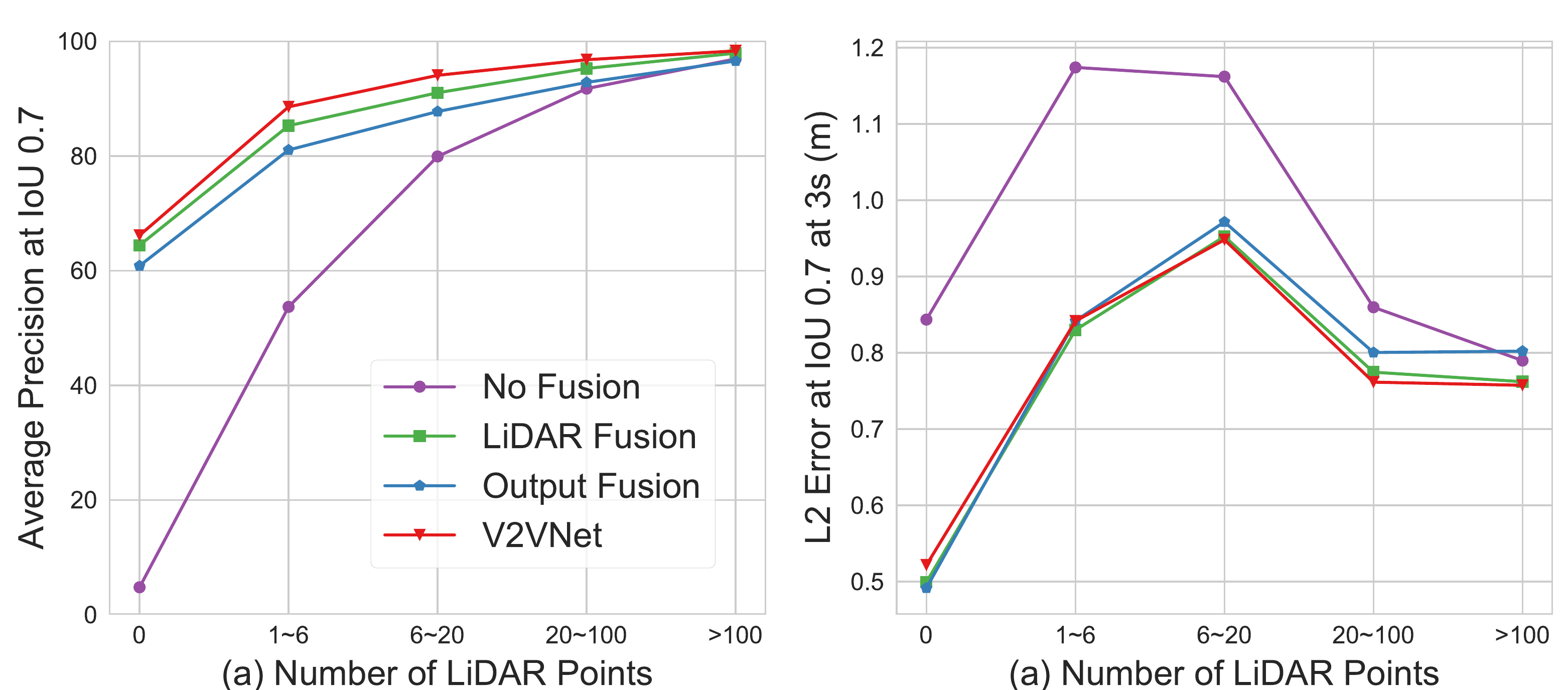}}
  \subfloat{\includegraphics[width=0.495\linewidth]{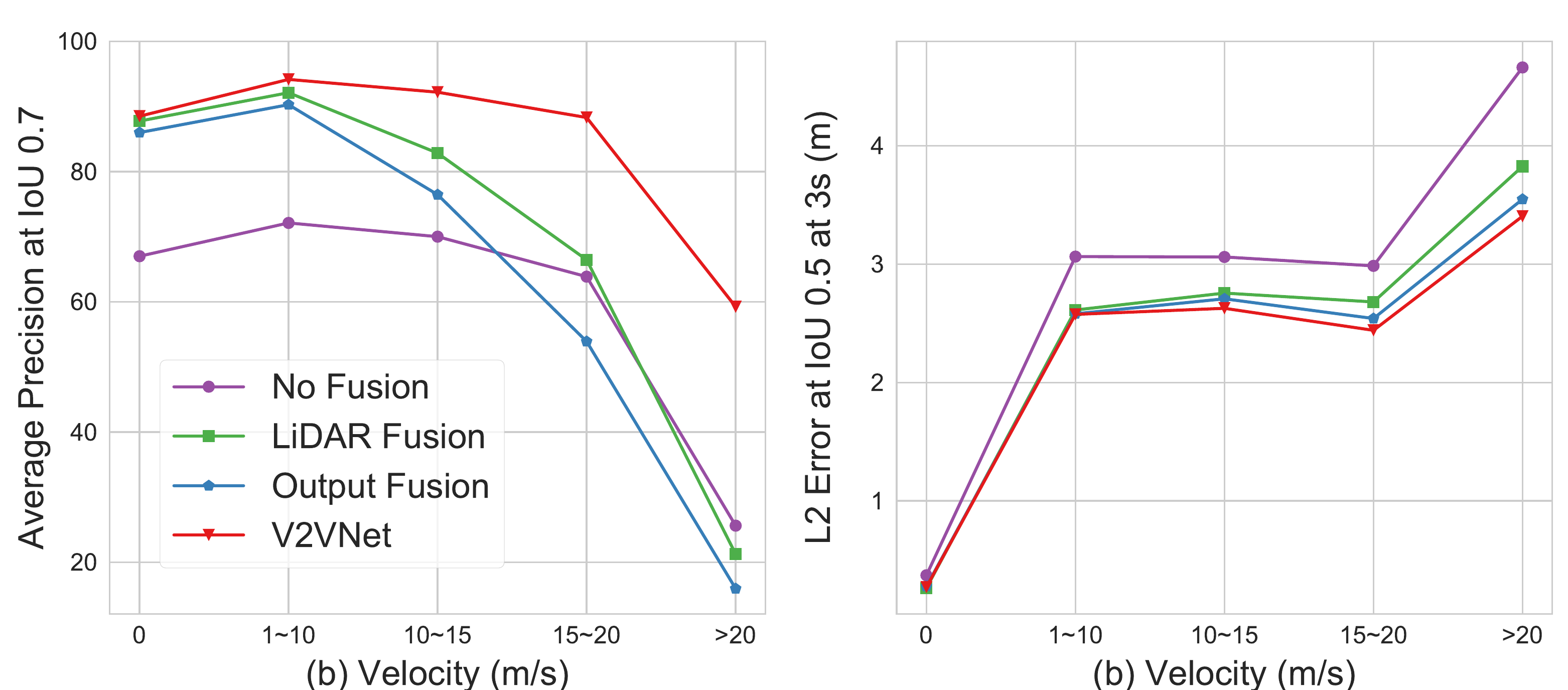}}
  \caption{Performance on objects with (first two columns) different number of LiDAR point observation (last two columns) different velocities.}
  \label{fig:diff_performance}
\end{figure*}

In this section we showcase the performance of our approach compared to other transmission and aggregation strategies as well as single vehicle P\&P.

\subsubsection{Metrics:}

We evaluate both detection and motion forecasting around the ego-vehicle with a range of: $x \in [-100, 100]$m, $y \in [-40, 40]$m. We include completely occluded objects ($0$ LiDAR points hit the object), making the task much more challenging and realistic than 
standard benchmarks.
For object detection, we compute Average Precision (AP) and Precision-Recall (PR) Curve at Intersection-over-Union (IoU) threshold of 0.7.
For motion forecasting, we compute absolute $\ell_2$ displacement error of the object center's location at future timestamps (3s prediction horizon with 0.5s interval) on true positive detections. We set the IoU threshold to 0.5 and recall to 0.9 (we pick the highest recall if 0.9 cannot be reached) to obtain the true positives. 
These values were chosen such that  we retrieve most objects, which is critical for safety in self-driving.
Note that most self-driving systems adopt this high recall as operating point.
We also compute Trajectory Collision Rate (TCR),  defined as the collision rate between the predicted trajectories of detected objects, where collision occurs when two cars overlap with each other more than a specific IoU (i.e., collision threshold $\tau$). This metric evaluates whether the predictions  are consistent with each other.
We exclude the other SDVs 
during evaluation, as those can be trivially predicted.

\subsubsection{Baselines:}
We evaluate the single vehicle setting, dubbed \textit{No Fusion}, which consists of LiDAR backbone network and output headers only, without V2V communication. We also  introduce two baselines for V2V communication: \textit{LiDAR Fusion} and \textit{Output Fusion}.  \textit{LiDAR Fusion} warps all received LiDAR sweeps from other vehicles to the coordinate frame of the receiver
via the relative transformation between vehicles (which is known, as all SDVs are assume to be localized) and performs direct aggregation.
We use the state-of-the-art LiDAR compression algorithm Draco~\cite{draco} to compress \textit{LiDAR Fusion} messages.
For \textit{Output Fusion}, each vehicle sends post-processed outputs, i.e., bounding boxes with confidence scores, and predicted future trajectories after non-maximum suppression (NMS). At the receiver end, all bounding boxes and future trajectories are first transformed to the ego-vehicle coordinate system and then aggregated across vehicles. NMS is then applied again to produce the final results.

\subsubsection{Experimental Details:}
For all analysis we set the maximum number of SDVs per scene to be 7 
(except for an ablation study measuring how the number of SDVs affect V2V performance in Fig. \ref{fig:num_veh}).
All models are trained with 
Adam 
\cite{adam}.

\subsubsection{Comparison to Existing Approaches:}
As shown in \tabref{tab:general},
V2V-based models
significantly outperform
\textit{No Fusion} on detection ($\sim$20\% at IoU 0.7) and prediction ($\sim$0.2 m $\ell_2$ error reduction at 3 sec.).  \textit{LiDAR Fusion} and \textit{V2VNet} also show strong reduction (20\% at 0.01 collision threshold) in TCR. These results demonstrate that all types of V2V communication 
provide substantial performance gains.
Among all V2V approaches, \textit{V2VNet} is either on-par with \textit{LiDAR Fusion} (which has no information loss) or achieves the best performance.
V2VNet's slight performance gain over \textit{LiDAR Fusion} may come from using the GNN in the cross-vehicle aggregation stage to reason about different vehicles' feature maps more intelligently than naive aggregation.
\textit{Output Fusion}'s drop in performance for TCR is due to the large number of false positives  relative to other V2V methods (see detection PR curve \figref{fig:pr_curve_and_l2_lambda}, left,  at recall $>$ 0.6).
\figref{fig:pr_curve_and_l2_lambda} shows the percentage of objects with an $\ell_2$ error at 3s smaller than a constant. %
This metric shows similar trends consistent with  \tabref{tab:general}.

\begin{figure*}[t]
  \centering
  \subfloat{\includegraphics[width=0.495\linewidth]{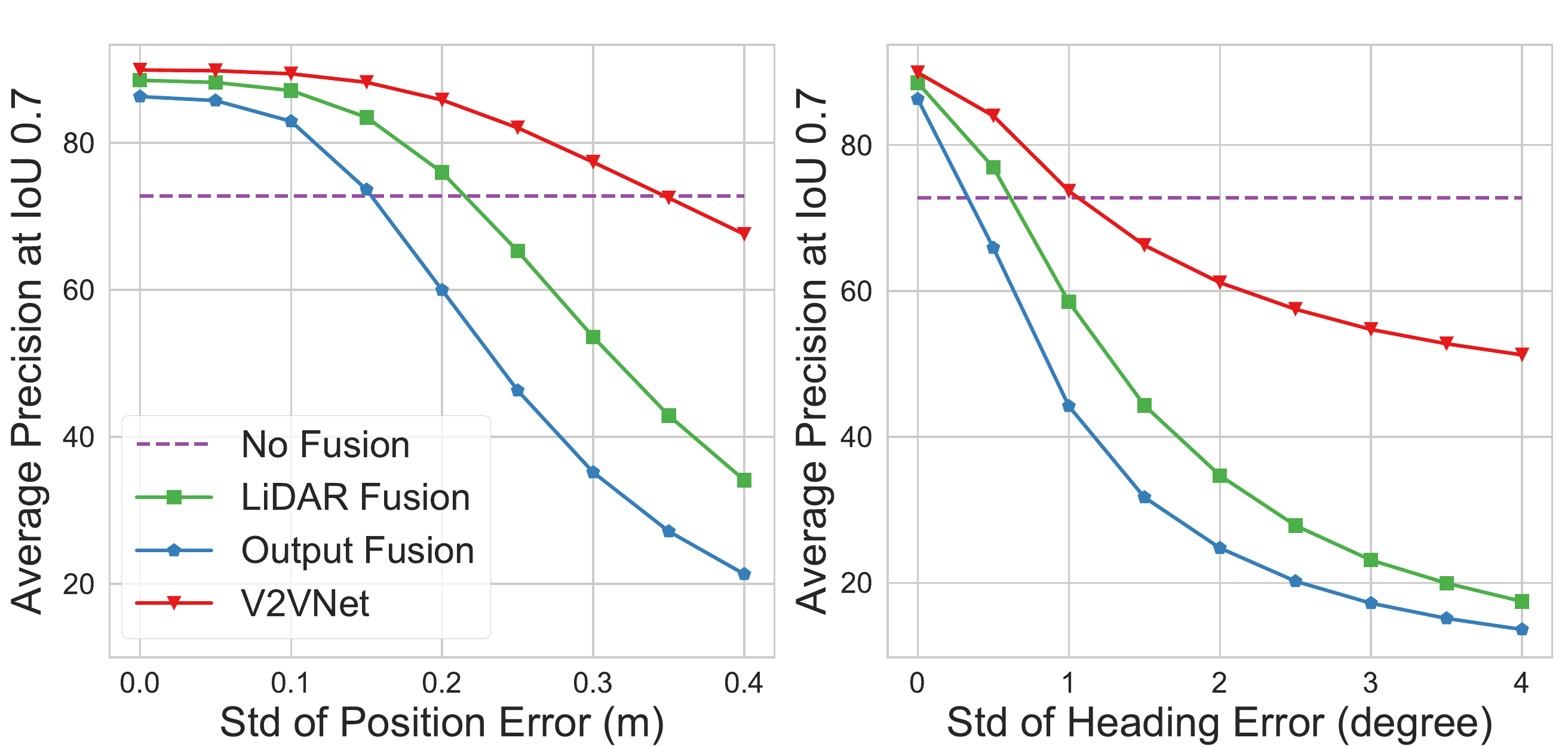}}
  \hfill
  \subfloat{\includegraphics[width=0.495\linewidth]{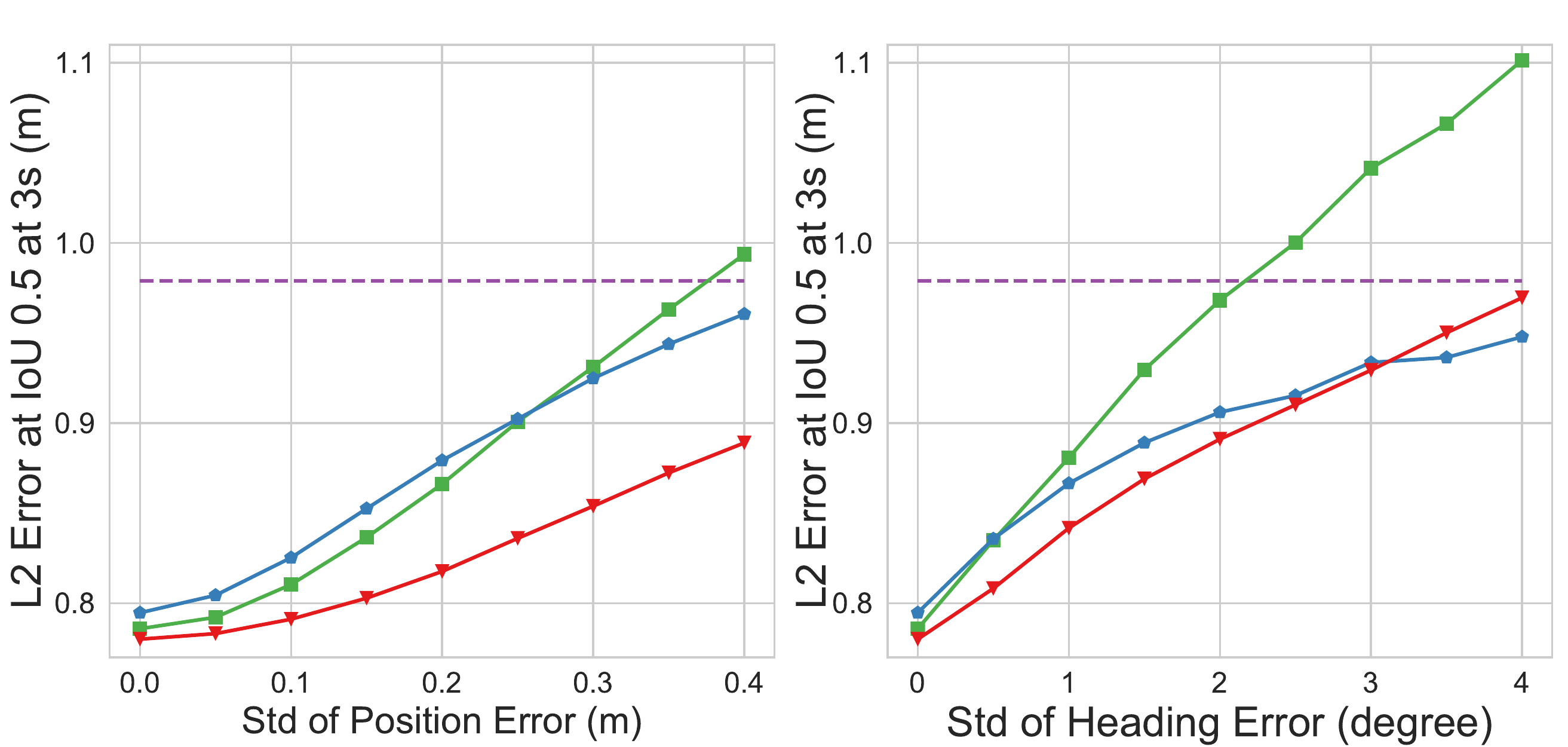}}
  \caption{Robustness on noisy vehicles' relative pose estimates.}
  \label{fig:pose_noise}
\end{figure*}

\subsubsection{Compression:}
\label{ssec:exp_comp}
\figref{fig:compression} shows the tradeoff between transmission bandwidth and accuracy for different V2V methods with and without compression.
Draco \cite{draco} achieves  33x  compression for \textit{LiDAR Fusion}, while our compressed intermediate representations achieved a 417x compression rate.
Note that compression marginally affects the performance.
This shows that the intermediate \pnp{}  representations
are much easier to compress than LiDAR.
Given the message size for one timestamp with a sensor capture rate of 10Hz, we compute the transmission delay based on V2V communication protocol~\cite{kenney2011dsrc}.
At the broadcast range 120 meters, the data rate is roughly 25 Mbps.
This means sending \textit{V2VNet} messages may induce roughly 9ms delay, which is very low.

\subsubsection{SDV Density:}
We now investigate how V2V performance changes as a function of \% of  SDVs in the scene.
To make this setting like the real world, for a given 25s snippet, we choose a fraction of candidate vehicles in the scene to be SDVs for the whole snippet.
As shown in \figref{fig:num_veh}, V2V
performance increases linearly with the \% of SDVs in both detection and prediction.

\subsubsection{Number of LiDAR points,  Velocity:}
As shown in  \figref{fig:diff_performance} (a) V2V methods boost the performance
on completely- and mostly-occluded objects (0 and 1$\sim$6 LiDAR points) by over 60\% in AP.
This is an extremely exciting result, since the main challenges of
perception and motion forecasting
are objects with very sparse observations.
 \figref{fig:diff_performance} (b) shows  performance on objects with different velocities.
While other V2V methods drop in detection performance as object velocity increases, \textit{V2VNet} has consistent performance gains over \textit{No Fusion} on fast moving objects.
\textit{Output Fusion} and \textit{LiDAR Fusion} may have deteriorated due to the rolling shutter of the moving SDV and the motion blur of moving agents during the temporal sweep of the LiDAR sensor.
These effects are more severe in the V2V setting, where SDVs may be moving in opposite directions at high speeds while recording moving actors.
Although not explicitly tackling such issue, \textit{V2VNet} performs contextual and iterative reasoning on information from different vehicles, which may indirectly handle   rolling shutter inconsistencies.

\begin{figure}[t]
  \centering
  \includegraphics[width=1.0\linewidth]{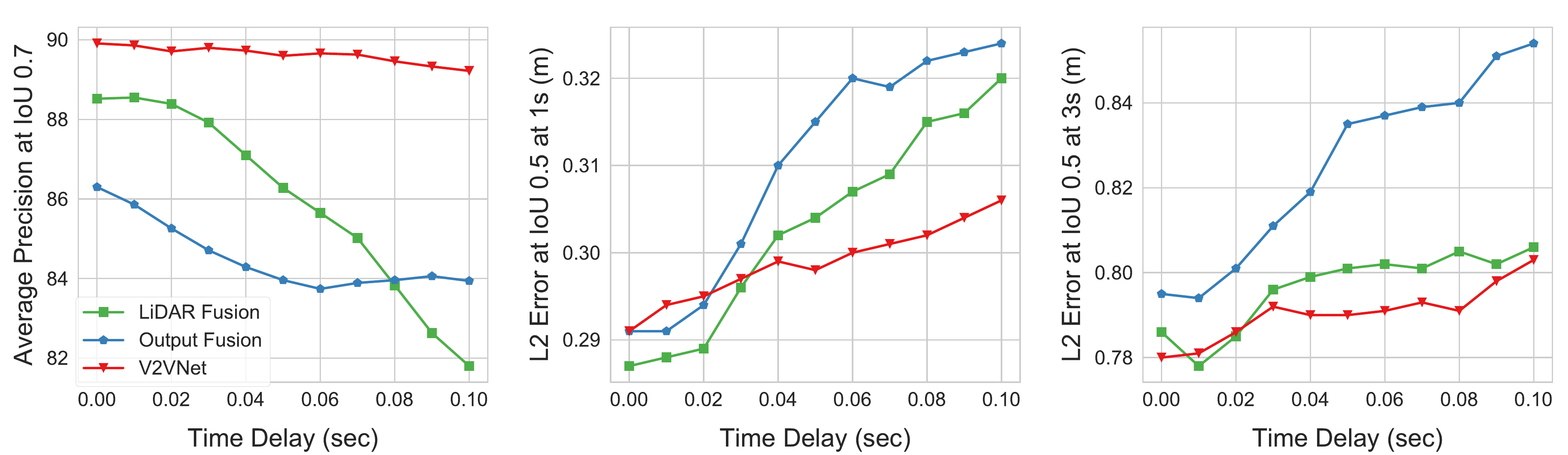}
  \caption{Effect of time delay in data exchange.}
  \label{fig:time_delay}
\end{figure}

\subsubsection{Imperfect Localization:}
 We simulate inaccurate pose estimates by introducing different levels of Gaussian ($\sigma = 0.4m$) and von Mises ($\sigma = 4 ^{\circ}$; $\frac{1}{\kappa}= 4.873\times 10^{-3}$) noise to position and heading of the transmitting SDVs. 
As shown in \figref{fig:pose_noise}, on both noise types, \textit{V2VNet} outperforms \textit{LiDAR Fusion} and \textit{Output Fusion} in
\pnp{}
performance. The only exception is \textit{Output Fusion} $\ell_2$ error with heading noise larger than $ 3 ^{\circ}$. We hypothesize that \textit{Output Fusion's} performance is better at this setting due to its low-recall (fewer true positives) relative to V2VNet (0.62 vs. 0.73 at $4 ^{\circ}$ noise).
Fewer true positives can cause lower $\ell_2$ error relative to 
higher recall methods.
Degradation from heading noise is more severe than position noise, as subtle rotation in the 
ego-view
will cause substantial misalignment for far-off objects;
a vehicle bounding box (5m x 2m) rotated by $1 ^{\circ}$ with respect to a pivot 70m away generates an IoU of 0.39 with the original.

\subsubsection{Asynchronous Propagation:}
We simulate random time delay by delaying the messages of other vehicles at random from $\mathcal{U}(0, t)$, where $t=0.1$. 
We apply a piece-wise linear velocity model (computed via finite differences) in \textit{Output Fusion} to compensate for time delay.
We do not make adjustments for \textit{LiDAR Fusion} as it is non-trivial. %
As shown in \figref{fig:time_delay},
\textit{V2VNet} demonstrates 
robustness across different time delays.
\textit{Output Fusion} does not perform well 
at high time delays as the piece-wise linear model used
is 
sensitive to 
velocity estimates.

\begin{figure*}[t]
  \centering
  \includegraphics[width=1.0\linewidth]{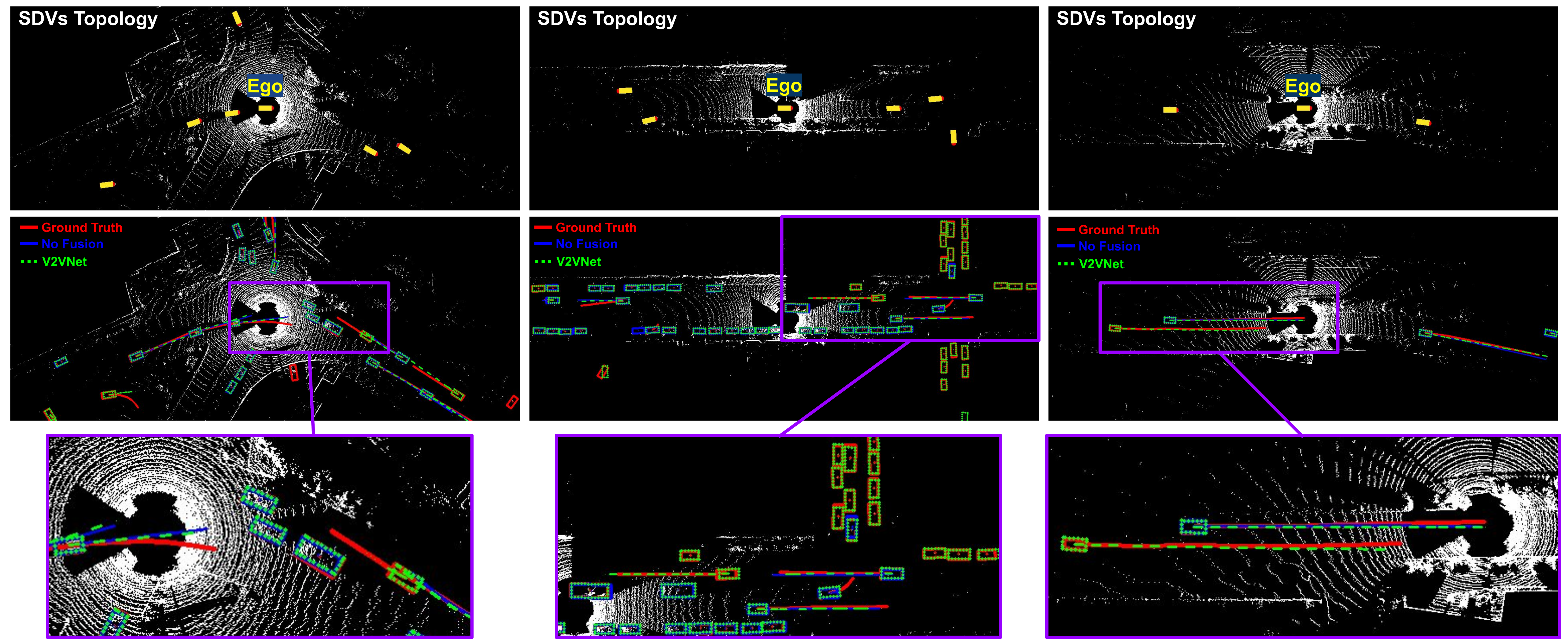}
  \caption{V2V-Net Qualitative Examples. Left: Occluded car detected; Middle:
Perception range increased; Right: Fast car detected.}
  \label{fig:qualitative}
\end{figure*}

\subsubsection{Mixed Fleet:}
We also investigate the case 
that the SDV
may receive different types of perception messages  (i.e., sensor data, intermediate representation and \pnp{} outputs). 
We analyze the setting where every SDV (other than the receiving vehicle) has 
1/3 %
chance to broadcast each measurement type.
We then perform \textit{Sensor Fusion}, \textit{V2VNet}, \textit{Output Fusion} for the relevant set of messages to generate the final 
output.  
The result is in between the 
three V2V
approaches: 
88.6 AP at IoU=0.7 for detection, 0.79 m error at 3.0s prediction, and 2.63 TCR.

\subsubsection{Qualitative Results:} As shown in Fig. \ref{fig:qualitative}, V2VNet can see 
further
and handle occlusion. 
For example, in Fig. \ref{fig:qualitative} far right, we perceive and motion forecast a high-speed vehicle in our right lane, which can give the downstream planning system more information to better plan a safe maneuver for a lane change. 
\textit{V2V-Net} also detects many more vehicles in the scene that were originally not detected by \textit{No Fusion} (Fig. \ref{fig:qualitative}, middle).

\section{Conclusion}

In this paper, we have proposed a V2V 
approach for perception and prediction that transmits compressed intermediate representations 
of the \pnp{} neural network,
achieving the best compromise between accuracy improvements and bandwidth requirements. 
To demonstrate the effectiveness of our approach we have created a novel {\it V2V-Sim} 
dataset that realistically simulates the world when SDVs will be ubiquitous.
We hope that our findings will inspire future work in V2V perception and motion forecasting strategies for safer self-driving cars.

%\subsubsection{Acknowledgments.} We gratefully acknowledge James Tu for valuable contributions  in the final paper. 

\bibliographystyle{splncs04}
\bibliography{egbib}
\end{document}